\newenvironment{myitem}{\begin{list}{$\bullet$}
{\setlength{\itemsep}{0pt}
\setlength{\topsep}{-5pt}
\setlength{\leftmargin}{12pt}
\setlength{\parsep}{0pt}
\setlength{\itemsep}{0pt}
\setlength{\partopsep}{0pt}}}%
{\end{list}}

\documentclass[letterpaper, 10 pt, conference]{ieeeconf}  

\usepackage{amsmath}
\usepackage{algorithm}
\usepackage{multicol}
\usepackage[bookmarks=true]{hyperref}
\usepackage{graphicx}
\usepackage[noend]{algpseudocode}
\usepackage{amssymb}
\usepackage{booktabs}

\IEEEoverridecommandlockouts                              

\title{\LARGE \bf
Model Predictive Control of Tensegrity Robots  \\ via Contact-Aware Graph Neural Dynamics Model
}

\author{%
Nelson Chen\textsuperscript{1},
Patrick Meng\textsuperscript{1},
Charles Tang\textsuperscript{1},
Angelina Degay\textsuperscript{1},
Zachary Brei\textsuperscript{2}, \\
Rebecca Kramer-Bottiglio\textsuperscript{2},
Kostas E. Bekris\textsuperscript{1},
Mridul Aanjaneya\textsuperscript{1}
\thanks{\textsuperscript{1}Computer Science, Rutgers University, Piscataway, NJ, USA. Email: \texttt{\{nelson.chen, patrick.meng, charles.tang, angelina.degay, kostas.berkris, mridul.aanjaneya\}@rutgers.edu}.}
\thanks{\textsuperscript{2}Mechanical Engineering, Yale University, New Haven, CT, USA. Email: \texttt{\{zachary.brei, rebecca.kramer\}@yale.edu}.}%
\thanks{}%
}

\begin{document}

\maketitle
\thispagestyle{empty}
\pagestyle{empty}

\begin{abstract}
Tensegrity robots offer lightweight, compliant mobility over challenging terrain but remain difficult to model and control due to complex contact-rich dynamics and partial observability. This work presents a model predictive path integral (MPPI) controller for a three-bar tensegrity robot driven by a learned graph neural network (GNN) dynamics model. This work first extends prior GNN-based models with a differentiable contact detection module. The extension allows the dynamics model to reason over non-horizontal planar terrains, obstacles, as well as self-collisions. Then, the learned dynamics model and the MPPI controller operate in a closed data-collection loop, iteratively improving model accuracy and control performance. This work further introduces a hybrid MPPI strategy that combines MPPI with turning motion primitives to improve maneuverability. Experiments are performed in MuJoCo across five navigation tasks, which include, wall obstacles, inclines, narrow corridors, low-clearance structures, and a composite 3D obstacle course. The experiments demonstrate that the hybrid MPPI controller operating over the learned GNN dynamics model improves predictive accuracy over a flat-ground baseline model and achieves superior navigation performance compared to $A^*$-based re-planning and MPPI-only variants. Results show that the contact-aware learned dynamics combined with the sampling-based model predictive control enable robust tensegrity navigation in complex, contact-rich environments.
\end{abstract}

\section{INTRODUCTION}

Tensegrity robots are an emerging class of hybrid rigid–soft mobile platforms designed for navigating uneven terrain and unstructured environments. Composed of rigid struts (rods) connected by flexible elements (cables), tensegrity structures are lightweight and low-cost while exhibiting high durability and adaptability~\cite{shah2022tensegrity}. However, the same cable–rod configuration that enables these advantages also introduces significant challenges for modeling and control. The dynamics of tensegrity systems involve strong coupling between compliant structural elements and complex contact interactions with the environment, making accurate dynamics modeling difficult.

\begin{figure}[t]
    \vspace{0.07in}
    \centering
    \includegraphics[width=0.9\linewidth]{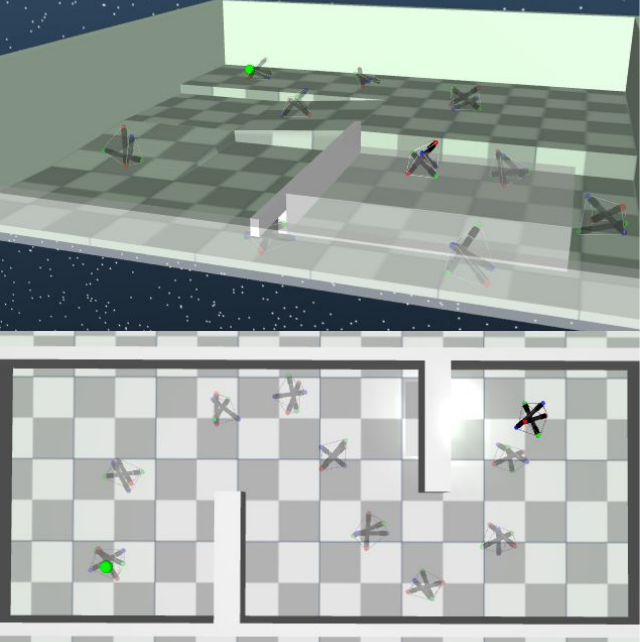}
    \vspace{-.1in}
    \caption{Two of the five obstacle courses used to demonstrate the MPPI controller's capabilities given the proposed GNN dynamics model. The non-transparent tensegrity represents the starting position, and the green sphere represents the goal. \textbf{TOP}: 3D obstacle course containing narrow turns and corridors, low clearance structure, and ramps. \textbf{BOTTOM}: Flat obstacle course with two walls that forces the robot to traverse an S-shaped path.}
    \vspace{-.26in}
    \label{fig:obs_course_1}
\end{figure}

These challenges have limited the development of sophisticated controllers for autonomous tensegrity locomotion. Prior work~\cite{vespignani2018steerable, baines2020rolling} has largely avoided the system dynamics by planning geometric solutions by tiling the tensegrity robot's polygonal faces. Other approaches~\cite{rl4tensegrity, zhang2025morphologyawaregraphreinforcementlearning} rely on reinforcement learning policies trained in simulation and transferred to real tensegrity platforms, but these policies have demonstrated limited locomotion capabilities. The difficulty of obtaining full state observations in physical systems further complicates the problem, often resulting in open-loop execution only. More recently,~\cite{Johnson2025Impact,our_ral} demonstrated closed-loop navigation through real-time re-planning over motion primitives. In parallel, accurately learned dynamics models for tensegrity robots based on graph neural networks (GNNs) operating under partial observations have been proposed~\cite{chen2026cablerobotgraphsimgraphneuralnetwork}, suggesting a promising pathway toward model-based control methods such as model predictive control (MPC).

\begin{figure*}[tb]
    \centering
    \vspace{0.07in}
    \includegraphics[width=\linewidth]{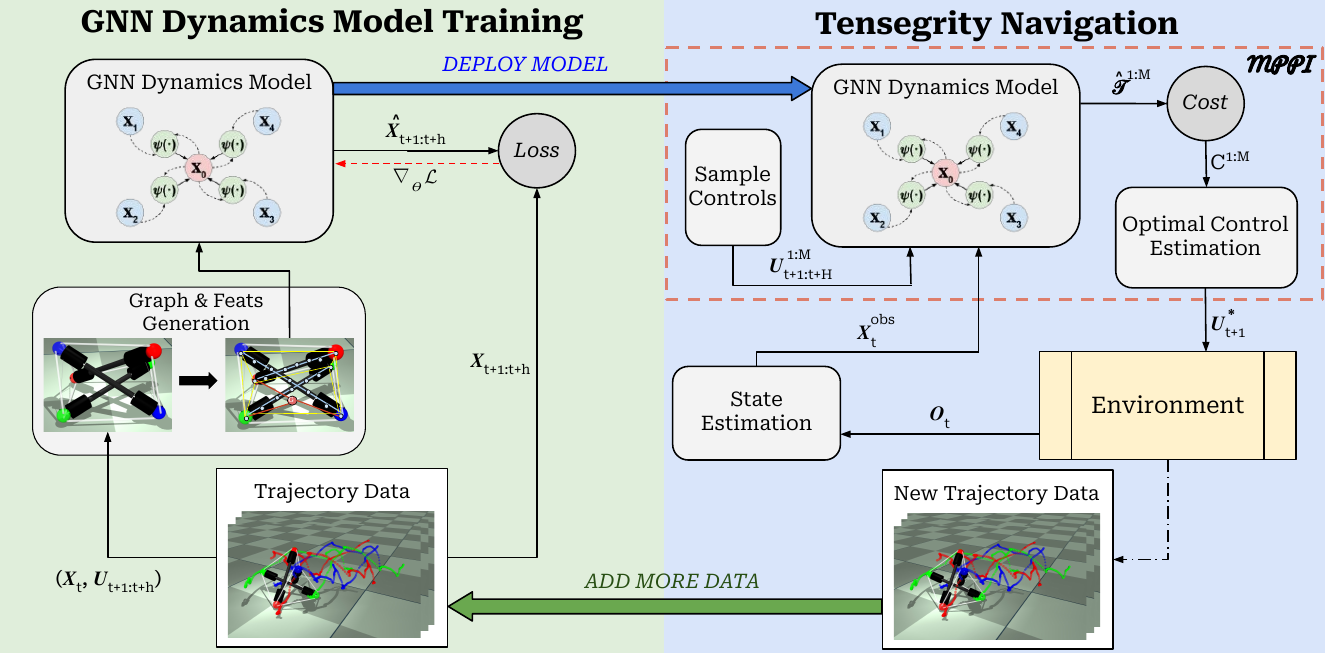}
    \vspace{-.25in}
    \caption{This figure highlights the GNN dynamics model training and MPPI data collection loop. \textbf{LEFT}: The GNN dynamics training process. The trajectory dataset provides a state $X_t$ and the next $h$ controls $U_{t+1:t+h}$. A graph and its features are generated and pushed to the GNN dynamics model for prediction. The model predicts the next $h$ states $\hat{X}_{t+1:t+h}$, which are compared against the ground-truth, $X_{t+1:t+h}$, to compute a loss and gradients with respect to the model parameters to train the model. After the modeling training has converged, it is deployed for tensegrity navigation, shown on the right diagram. \textbf{RIGHT}: The tensegrity navigation process using an MPPI controller. $M$ sampled control sequences of horizon $H$, $U^{1:M}_{t+1:t+H}$ and the current estimated state are passed to the GNN. The GNN then autoregressively simulates the trajectories $\mathcal{T}^{1:M}$ in parallel and their costs $\mathcal{C}^{1:M}$. The next optimal control $U^{*}_{t+1}$ is estimated by the MPPI algorithm and passed to the environment to be executed. This closed-loop is run until the desired task is complete or the time limit expires. These trajectories are collected and added to the training data to train the next iteration of the GNN dynamics model.}
    \vspace{-.23in}
    \label{fig:pipeline}
\end{figure*}

This work develops an MPC framework for tensegrity locomotion based on a learned GNN dynamics model with explicit contact modeling. The dynamics model incorporates a differentiable contact detection module that generates contact features representing interactions with environmental planar surfaces and other structural elements, as well as self-collisions within the tensegrity structure. The learned contact representations are embedded as node and edge features in the graph-based dynamics model, allowing the system to capture contact-dependent dynamics necessary for predictive control. A model predictive path integral (MPPI) controller uses this learned dynamics model as its internal predictive model, enabling closed-loop planning and control in environments with complex contact interactions.

To evaluate the proposed approach, experiments assess both the learned dynamics model and the resulting MPPI controller. The extended GNN dynamics model is first compared against a baseline GNN model that assumes flat-ground contacts only, and an ablation study evaluates the impact of the contact features and edges. The MPC controller is then evaluated in MuJoCo across several navigation scenarios: (i) a flat obstacle course, (ii) a flat narrow corridor, (iii) a long incline ramp, (iv) a low-clearance structure, and (v) a three-dimensional obstacle course combining elements from the previous scenarios. The GNN dynamics model is only trained on data collected from courses (i) to (iv), hence the final course serves as an unseen environment for the GNN model. Performance is measured using average success rate and task completion time, comparing a baseline $A^*$ primitive re-planning approach, MPPI-only controller, and a hybrid controller combining MPPI with turning motion primitives.

The contributions of this work are summarized as follows:

\begin{myitem}
\item A learned GNN dynamics model with differentiable contact detection that extends prior efforts to also capture interactions with non-horizontal planar terrains, environmental obstacles, and self-collisions.
\item An MPC framework for tensegrity robots using MPPI control over the learned GNN dynamics model leads to an iterative data collection loop that allows improving the dynamics model and the abilities of the controller.
\item A hybrid MPPI and turning primitives solution for more robust maneuverability.
\item Experimental evaluation in simulation of the proposed dynamics model and MPPI controller across diverse terrain configurations and navigation tasks.
\end{myitem}

\section{RELATED WORKS}

Early work on tensegrity robot locomotion largely avoided system dynamics and instead relied on geometric planning strategies. These approaches used face transition rules, where the polygonal faces of a tensegrity structure are tiled to determine feasible rolling transitions for navigation~\cite{vespignani2018steerable,baines2020rolling}. Later work incorporated dynamics during the offline design phase while still avoiding runtime dynamics reasoning. One such line of work manually designs locomotion primitives and precomputes their translational and rotational effects using system-identified simulators for use in higher-level planning~\cite{our_ral, Johnson2025Impact}. Another line trains reinforcement learning policies in simulation and transfers them to physical tensegrity robots~\cite{rl4tensegrity,zhang2025morphologyawaregraphreinforcementlearning}. However, these approaches generally demonstrate limited locomotion capabilities such as reproducing behaviors similar to the human-engineered primitives.

Model predictive control (MPC) enables reasoning over system dynamics during execution by repeatedly optimizing control sequences using an internal dynamics model in a closed-loop manner~\cite{mpc_survey}. Despite its success in robotics, MPC has seen limited application in tensegrity systems. Existing work applies MPC only to a spine tensegrity module that serves as the backbone of a quadruped platform~\cite{sabelhaus2018trajectorytrackingcontrolflexible, sabelhaus2019modelpredictivecontrolinversestatics}. Applying MPC to free-rolling tensegrity robots remains challenging due to their highly nonlinear dynamics and complex contact interactions. Model predictive path integral (MPPI) control~\cite{mppi} provides a sampling-based variant of MPC that evaluates candidate control sequences through forward rollouts of the dynamics model, enabling efficient parallel computation on GPUs and improving robustness to local minima.

Accurate dynamics models for tensegrity robots remain difficult to obtain. Early approaches relied on analytical simulators that capture tensegrity physics but are typically non-differentiable and sensitive to modeling errors~\cite{motes, Paul2006DesignAC, ntrt}. Differentiable simulators were later developed to improve system identification and reduce simulation-to-reality gaps~\cite{recurrent, r2s2r, sim2sim}. However, these methods generally assume full observability of the system state, limiting their applicability for runtime control in real-world settings.

More recently, learning-based approaches have modeled tensegrity dynamics directly from data. Graph neural network (GNN) models leverage the natural graph structure of cable-driven systems and enable dynamics learning under partial observations~\cite{chen2024learningdifferentiabletensegritydynamics, chen2026cablerobotgraphsimgraphneuralnetwork}. Existing models, however, assume horizontal-ground contacts and do not explicitly represent interactions with non-horizontal surfaces or self-collisions. This work extends GNN-based dynamics models with differentiable contact detection to capture interactions with non-horizontal planar terrains, workspace obstacles, and internal self-collisions. The resulting contact-aware dynamics model is integrated within an MPPI controller to enable model predictive control for tensegrity navigation.

\begin{figure}[t]
    \centering    
    \vspace{0.07in}
    \includegraphics[width=0.7\linewidth]{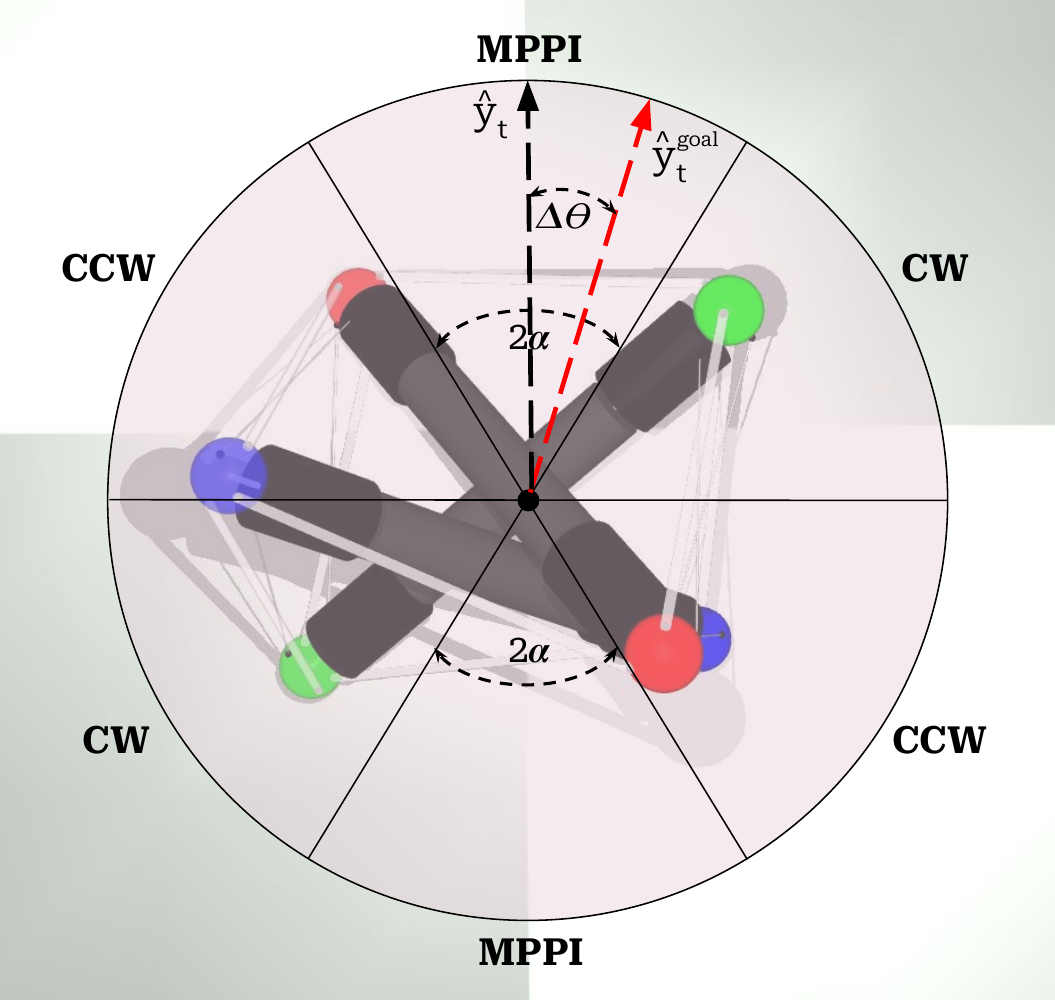}
    \caption{Hybrid MPPI and turning motion primitives strategy. Given the current robot heading $\hat{\mathsf{y}}_t$ and a user-defined threshold $\alpha$, the robot's angular domain relative to $\hat{\mathsf{y}}_t$ is divided up into regions for MPPI, clockwise (cw) turn primitive, and counter-clockwise (ccw) turn primitive. If the angular difference $\Delta \theta$ between the estimated cost gradient $\hat{\mathsf{y}}^{goal}_t$ and ($\hat{\mathsf{y}}_t$ or $-\hat{\mathsf{y}}_t$), then the MPPI controller is executed. If not, then turn motion primitives are executed to try and align the robot into the MPPI regions. For that reason, if $\Delta \theta$ is between $\alpha$ and $\frac{\pi}{2}$, then a clockwise turn is executed, otherwise a counter-clockwise turn is executed.}
    \vspace{-.25in}
    \label{fig:hybrid_mppi}
\end{figure}

\section{ROBOT: THREE BAR TENSEGRITY}

The robot used in this work is a three-bar prismatic tensegrity based on the open-source design described in~\cite{our_ral}, shown in Fig.~\ref{fig:obs_course_1}. The structure consists of three identical rods, each 33 cm in length. A colored end cap is attached to each rod to visually distinguish them during perception and tracking. The robot is actuated by two motors per rod that control cable lengths, enabling extension and contraction of the tensegrity structure. By adjusting these cable lengths, the robot changes its shape and shifts its center of mass (CoM), producing locomotion.

Each motor contains an encoder used to estimate the corresponding cable length. To further improve state estimation, capacitive strain sensors are embedded along the robot’s edges~\cite{johnson2022sensor}. The six short cables (three on each side of the structure) are actively actuated by the motors and run parallel to the strain sensors, while the three longer central tendons act as passive elastic elements and sensing cables. Together, these sensors provide partial observations of the robot configuration that are used for state estimation.

Several robot attributes used throughout this paper are defined next. The robot state is defined as the union of the individual rod states $X^{robot}_t := \{X^i_t \mid i=1,2,3\}$, where the state of rod $i$ at time $t$ is
$X^i_t = \{P^i_t, R^i_t, V^i_t,\Omega^i_t\}$. Here, $P^i_t$ denotes the position, $R^i_t$ the orientation, $V^i_t$ the linear velocity, and $\Omega^i_t$ the angular velocity. Each rod also has two endpoints $(e^{i,\text{left}}_t, e^{i,\text{right}}_t)$. The rod’s principal axis $\hat{r}^i_t$ is defined as the unit vector pointing from the left endpoint to the right endpoint.

With these quantities, the robot center of mass is defined as the mean of the rod positions. Similarly, the robot principal axis is defined as the normalized mean of the rod principal axes:

\vspace{-0.26in}
\begin{align}
P^{robot}_t = \frac{1}{3}\sum P^i_t\;,\;\;\; \hat{r}^{robot}_t = \frac{1}{3}\sum \hat{r}^i_t
\end{align}

Although the robot operates in three-dimensional space, the full 3D dynamics model is only needed for prediction. The navigation cost function is, however, formulated in two dimensions to simplify goal-directed navigation. Accordingly, the robot’s planar principal axis $\hat{\mathsf{x}}_t$ is defined as the normalized projection of $\hat{r}^{robot}_t$ onto the $xy$-plane. The heading direction $\hat{\mathsf{y}}_t$ is then defined as a $90^\circ$ rotation of $\hat{\mathsf{x}}_t$ about the $\hat{z}$ axis. The resulting planar pose of the robot in $SE(2)$ is given by $(P^{robot}_{t,x}, P^{robot}_{t,y}, \theta_t)$, where $\theta_t$ denotes the robot heading angle.

\begin{algorithm}[h]
    \caption{Hybrid MPPI with Turning Motion Primitives}
    \label{algo:hybrid_mppi}
    \begin{algorithmic}[1]
    
        \State \textbf{Input:} Robot state $X^{robot}_t$, cost function $\mathcal{C}(\cdot)$, neighborhood radius $R$, orientation tolerance $\alpha$, turning primitives $\{cw, ccw\}$
        \State \textbf{Output:} ControlType $\in \{\texttt{MPPI}, \texttt{Primitives}\}$, actions
        
        \State Compute $SE(2)$ pose $(P^{robot}_{t,x}, P^{robot}_{t,y}, \theta_t)$ and heading $\hat{\mathsf{y}}_t$ from $X^{robot}_t$
        
        \State Compute neighborhood costs
        \State $\mathcal{C}^R = \{\mathcal{C}(x,y) | x \in (P^{robot}_{t,x}-R,P^{robot}_{t,x}+R), y\in(P^{robot}_{t,y}-R,P^{robot}_{t,y}+R)\}$
        
        \State Find minimum-cost point in neighborhood
        \State $(x^R_{min}, y^R_{min}) = \arg\min \mathcal{C}^R$
        
        \State Estimate local cost gradient direction
        \State $\hat{\mathsf{y}}^{goal} = (x^R_{min}-P^{robot}_{t,x},\; y^R_{min}-P^{robot}_{t,y})/\| \cdot \|_2$
        
        \State Compute signed angular differences
        \State $\Delta\theta_{forward} = \operatorname{atan2}(\hat{\mathsf{y}}_t \times \hat{\mathsf{y}}^{goal},\; \hat{\mathsf{y}}_t \cdot \hat{\mathsf{y}}^{goal})$
        \State $\Delta\theta_{backward} = \operatorname{atan2}(-\hat{\mathsf{y}}_t \times \hat{\mathsf{y}}^{goal},\; -\hat{\mathsf{y}}_t \cdot \hat{\mathsf{y}}^{goal})$
        \State $\Delta\theta = \min(\Delta\theta_{forward}, \Delta\theta_{backward})$
        
        \If{$|\Delta\theta| \leq \alpha$}
            \State ControlType $\gets$ \texttt{MPPI}
            \State actions $\gets$ MPPI$(X^{robot}_t)$
        \Else
            \State ControlType $\gets$ \texttt{Primitives}
            \If{$\alpha < \Delta\theta \leq \frac{\pi}{2}$}
                \State actions $\gets \{ccw\}$
            \Else
                \State actions $\gets \{cw\}$
            \EndIf
        \EndIf
        
        \State \Return ControlType, actions
    
    \end{algorithmic}
\end{algorithm}

\section{APPROACH}

\subsection{Model Predictive Path Integral Controller}

This work employs a model predictive path integral (MPPI) controller using an extended GNN dynamics model. MPPI is a sampling-based model predictive control method for stochastic optimal control. At each control step, MPPI samples $M$ candidate control sequences $\{U^m\}$ with horizon $H$. Given a dynamics model $f(\cdot)$, a cost function $\mathcal{C}(\cdot)$, and an initial state $X_0$, the sampled control sequences are rolled out through the dynamics model to generate $M$ trajectories $\{\mathcal{T}^m\}$ and their associated costs.

\vspace{-0.17in}
\begin{align}
\mathcal{T}^m = \Big[X_0, f(X_0,U^m_0), \ldots, f(X_{H-1},U^m_{H-1})\Big]
\end{align}

The optimal control sequence $U^*$ is estimated as a weighted average over the sampled control sequences using importance sampling weights $w^m$:

\vspace{-0.17in}
\begin{align}
w^m = \frac{1}{\eta}\exp\!\left(-\frac{1}{\beta}(\mathcal{C}(\mathcal{T}^m)-\rho)\right), 
\qquad \sum w^m = 1
\end{align}

\vspace{-0.17in}
\begin{align}
U^* = \sum w^m U^m
\end{align}

\noindent where $\eta$ is a normalization constant, $\beta$ is the inverse temperature parameter, and $\rho=\min_m \mathcal{C}(\mathcal{T}^m)$ is the minimum trajectory cost among the samples. This process is illustrated on the right side of Fig. \ref{fig:pipeline}.

\subsection{Hybrid MPPI and turning motion primitives} 
MPPI alone is often effective when the robot is sufficiently dexterous and near-optimal control sequences can be reliably sampled at runtime. For the three-bar tensegrity robot, however, the MPPI controller can readily discover forward and backward locomotion but struggles to sample effective turning behaviors. Discovering such motions typically requires long horizons and a large number of samples, making MPPI here prohibitively expensive. To address this limitation, a hybrid controller combining MPPI with human-engineered turning motion primitives is introduced. The key idea, as shown in Fig. \ref{fig:hybrid_mppi}, is to use MPPI when the robot’s heading direction (or its opposite) is aligned with the local gradient of the cost function, allowing the controller to exploit forward or backward motion. When the heading is misaligned with the cost gradient, the turning primitives are executed to reorient the robot toward the desired direction. Once alignment is established, MPPI resumes control for goal-directed locomotion. The full procedure is detailed in algorithm~\ref{algo:hybrid_mppi}.

\subsection{Obstacle-Aware, Wave-Front Cost Function}

A naive cost function is the Euclidean distance to the goal. Although computationally efficient and straightforward to evaluate, this metric can be misleading in cluttered environments because it ignores obstacle constraints. As a result, it may favor straight-line trajectories that intersect obstacles and are therefore infeasible, leading to overly optimistic estimates of the true cost-to-go in constrained workspaces.

To overcome this limitation, the workspace is discretized into a uniform grid and only collision-free grid cells are added as nodes to a graph. Edges are formed between adjacent collision-free grid cells. A search to the goal is performed over this graph. Efficiency is improved by reusing a shared closed list across solves, allowing previously expanded nodes with known optimal cost-to-go values to be reused so that subsequent searches terminate early while propagating obstacle-aware distance estimates outward from the goal. The steps to compute this cost function is shown in algorithm \ref{algo:wavefront_cost}.

\begin{algorithm}[t]
\caption{Wave-Front Cost Map Construction}
\label{algo:wavefront_cost}
\begin{algorithmic}[1]

\State \textbf{Input:} Goal state $\mathbf{g}$, boundary 
$\mathbf{B}=[x_{\min},x_{\max},y_{\min},y_{\max}]$, 
obstacle set $\mathcal{O}$, grid resolution $\Delta$

\State \textbf{Output:} Cost map $\mathbf{M}$

\State $\mathbf{M} \gets \varnothing$ \Comment{Hash map of costs}
\State $\mathcal{S} \gets \varnothing$ \Comment{Valid grid states}
\State $\mathcal{Q} \gets \{(\mathbf{g},0)\}$ \Comment{Priority queue}

\For{$x \in [x_{\min},x_{\max}]$ with step $\Delta$}
    \For{$y \in [y_{\min},y_{\max}]$ with step $\Delta$}
        \If{\textsc{CollisionDetect}$( (x,y), \mathbf{B}, \mathcal{O})$}
            \State $\mathbf{M}[(x,y)] \gets \infty$
        \Else
            \State $\mathcal{S} \gets \mathcal{S} \cup \{(x,y)\}$
        \EndIf
    \EndFor
\EndFor

\While{$\mathcal{S} \neq \varnothing$ \textbf{and} $\mathcal{Q} \neq \varnothing$}

    \State $(\mathbf{s}, c) \gets$ \textsc{PopMin}$(\mathcal{Q})$
    \State $\mathbf{M}[\mathbf{s}] \gets c$

    \For{neighbor $\mathbf{n} \in \mathcal{N}(\mathbf{s})$}
        \If{$\mathbf{n} \notin \mathbf{M}$}
            \State $\mathcal{Q} \gets 
            \mathcal{Q} \cup \{(\mathbf{n}, c + distance(\mathbf{s},\mathbf{n}))\}$
        \EndIf
    \EndFor

\EndWhile

\State \Return $\mathbf{M}$

\end{algorithmic}
\end{algorithm}

\subsection{Extended GNN Model for Planar Terrains and Obstacles}

\begin{figure}[tb]
    \centering
    \vspace{0.1in}\includegraphics[width=0.55\linewidth]{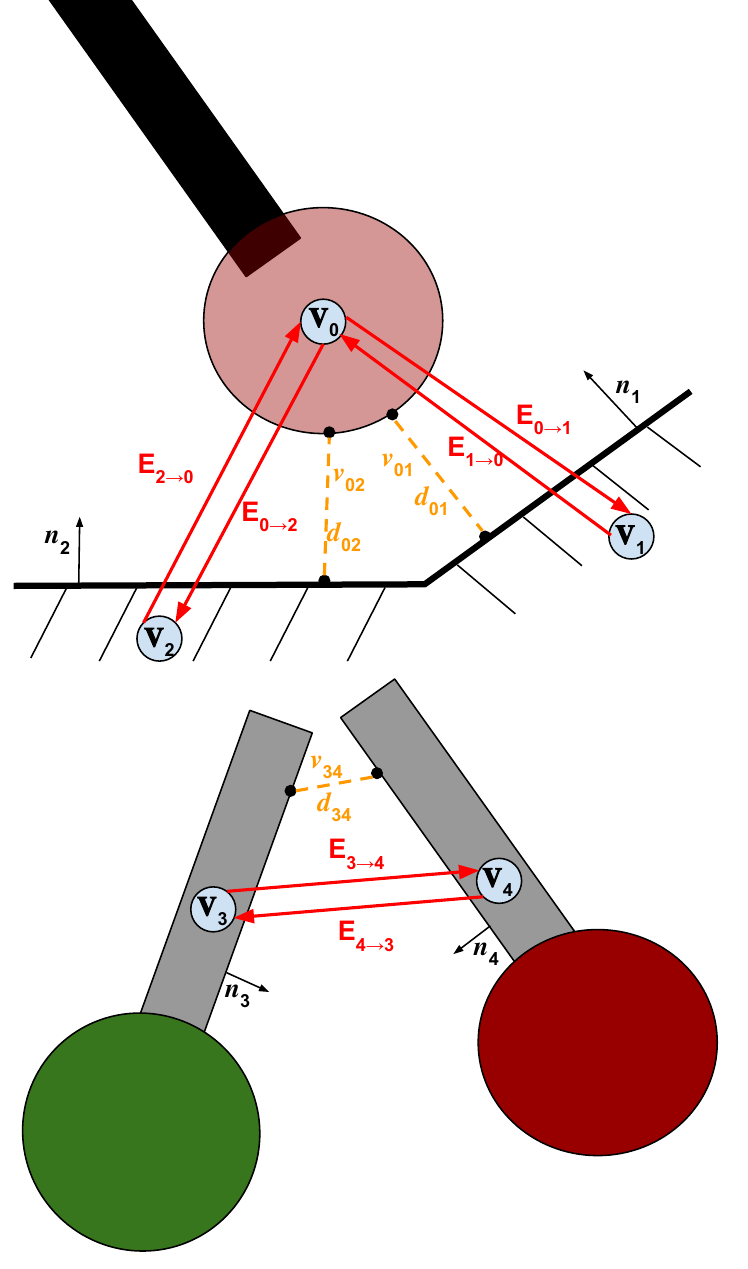}
    \vspace{-.1in}
    \caption{Collision feature generation. \textbf{TOP}: End cap to planar environment surfaces. \textbf{BOTTOM}: Self-collision between rods. For both environment collisions and self-collisions, nodes $(V_i,V_j)$ represent physical objects with surfaces. Given their states and geometries, the minimum signed distance $d_{ij}$ and the corresponding surface points are computed. At these points, relative velocity $v_{ij}$ and surface normals $(n_i,n_j)$ are calculated. The distance $d_{ij}$ is added as a node feature, while $(d_{ij}, n_{ij}, v_{ij})$ are encoded as edge features in $E^{con}_{sender\rightarrow receiver}$.}
    \vspace{-.2in}
    \label{fig:collision}
\end{figure}

The GNN dynamics model used in this work builds on \cite{chen2026cablerobotgraphsimgraphneuralnetwork} and is extended to handle non-horizontal planar terrains, environmental obstacles, and self-collisions between rods. The model consists of two stages: (i) graph and feature generation and (ii) graph neural processing, shown on the left side of Fig. \ref{fig:pipeline}.

In the graph generation stage, the robot state is converted into a graph representation. The graph contains body nodes (including environment and obstacle nodes) $\mathcal{V}_t=\{V_t^i\}$ with node states $x_t^i=(p_t^i,v_t^i)$, where $p_t^i$ is the position, and $v_t^i$ is the linear velocity. Nodes are connected by three edge types: body edges $\mathcal{E}_t^{body}=\{E^{body}\}$, cable edges $\mathcal{E}_t^{cable}=\{E^{cable}\}$, and contact edges $\mathcal{E}_t^{con}=\{E^{con}\}$. Node and edge feature vectors are computed from the physical state and properties of the represented entities. In the graph neural processing stage, an encode-process-decode architecture~\cite{fignet,gnn_rigid_body_sim} maps raw features into a latent space, performs message passing between nodes, and decodes predictions of the next $h$ changes in rod velocities $\Delta v^i_{t+1:t+h}$ and cable rest lengths $\Delta L^{rest,ij}_{t+1:t+h}$. These predictions are applied autoregressively for $H/h$ steps to simulate an MPPI rollout. Additional details of the base dynamics model are provided in \cite{chen2026cablerobotgraphsimgraphneuralnetwork}.

The base GNN model assumes contact only with a flat ground plane at $z=0$. To enable navigation over uneven planar environments, this work introduces a differentiable and parallelized contact detection module implemented in PyTorch. The module computes contact features and edges between the robot and planar environment objects, as well as self-collisions between cylindrical rods, and is integrated into the graph and feature generation stage.

At each graph generation step, the module receives the robot state and a set of planar environment objects. For robot–environment interactions, each end-cap node computes the minimum signed distance to nearby planar surfaces along with the corresponding surface points. From these points, surface normals and relative velocities are calculated. Signed distances are upper bound by a user-defined threshold to restrict interactions to local contacts only. For each node, the $k$ smallest distances are added as node features. Contact edges are dynamically assigned between end-cap nodes and planar surfaces if their distance falls below the threshold, with edge features containing the computed contact quantities. Self-collisions are handled using the same procedure between pairs of rods and their internal cylindrical nodes, allowing the model to capture rod–rod interactions during locomotion shown in Fig. \ref{fig:collision}.

\subsection{GNN-MPPI Data Collection Loop}

The extended GNN dynamics model and MPPI controller are combined in a loop to iteratively improve both the GNN and MPPI controller. As shown in Fig. \ref{fig:pipeline}, on one side, it is the model training phase, where a GNN dynamics model is trained using the current dataset that is bootstrapped with data from executing only primitives. The model is then deployed to the navigation phase where the MPPI controller is executed on the desired tasks, and new and diverse data is collected to help the model improve further in the next model training iteration. This can be run until the model and MPPI no longer improve.

\section{SIMULATION EXPERIMENTS \& RESULTS}

\subsection{GNN Dynamics Evaluation}

\begin{figure}[h]
    \centering
    \vspace{-.15in}  
    \includegraphics[width=0.3\textwidth]{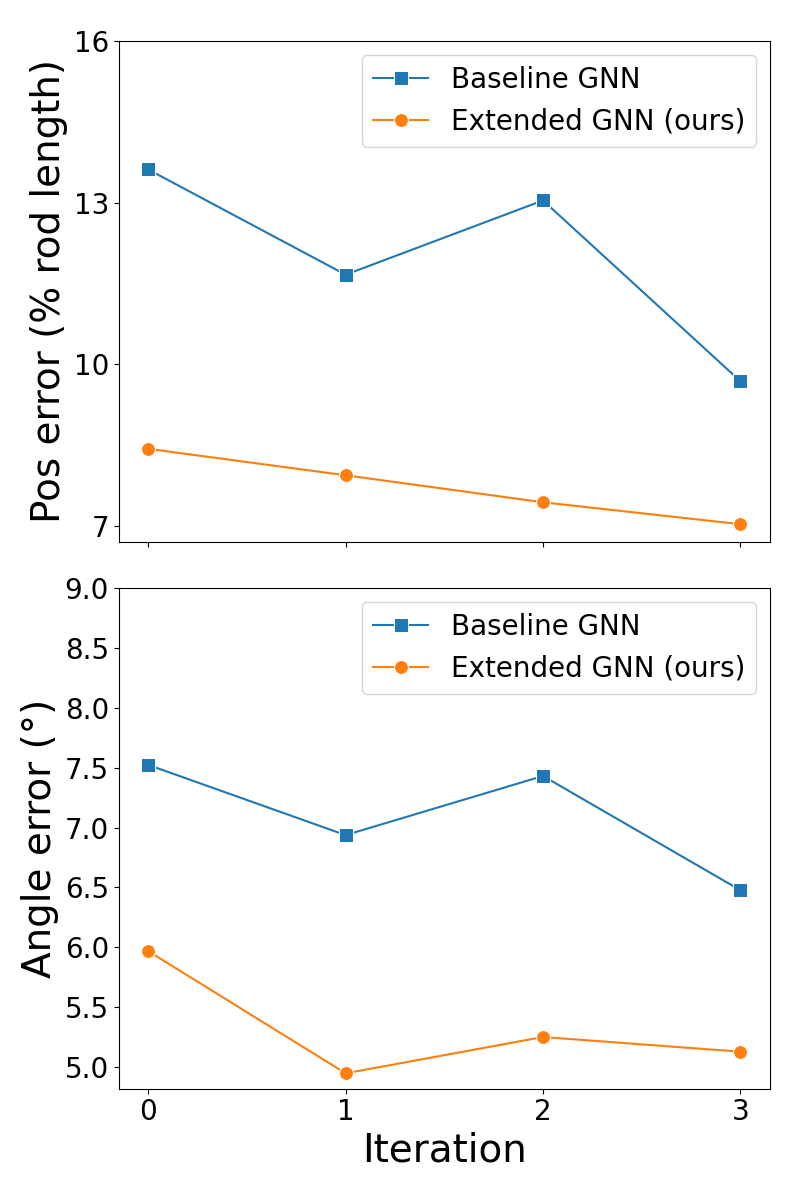}
    \vspace{-0.2in}
    \caption{Comparison of model errors between baseline GNN, which only assumes horizontal ground, across 4 GNN training and data collection iterations. \textbf{TOP}: Average mean-squared error of center-of-mass over the MPPI horizon of 2 seconds, normalized by the rod's length. \textbf{BOTTOM}: Average absolute angular error of the rods' principal-axes over the MPPI horizon of 2 seconds.}
    \vspace{-0.05in}
    \label{fig:model_eval_iteration}
\end{figure}

\begin{figure*}[tb]
    \centering
    \includegraphics[width=1.0\linewidth]{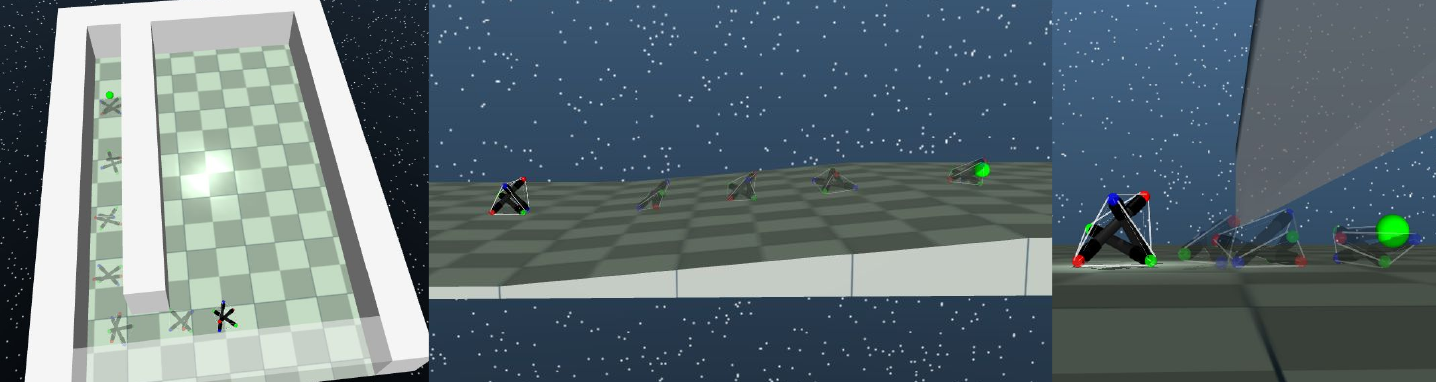}
        \vspace{-.2in}
    \caption{Three additional MPPI evaluation tasks/courses beyond those shown in Fig. \ref{fig:obs_course_1}. The goal is indicated by a green sphere in each course. \textbf{LEFT}: Narrow corner and corridor that is only slightly wider than the robot and requires careful alignment of the robot with the environmental constraints. \textbf{MIDDLE}: Long $5^\circ$ incline. \textbf{RIGHT}: Low clearance structure that introduces an obstacle below the resting height of the robot and forces it to significantly deform in order to successfully navigate the course.}
    \vspace{-.1in}
    \label{fig:obs_courses_2}
\end{figure*}

\begin{figure}[h]
    \centering
    \includegraphics[width=0.33\textwidth]{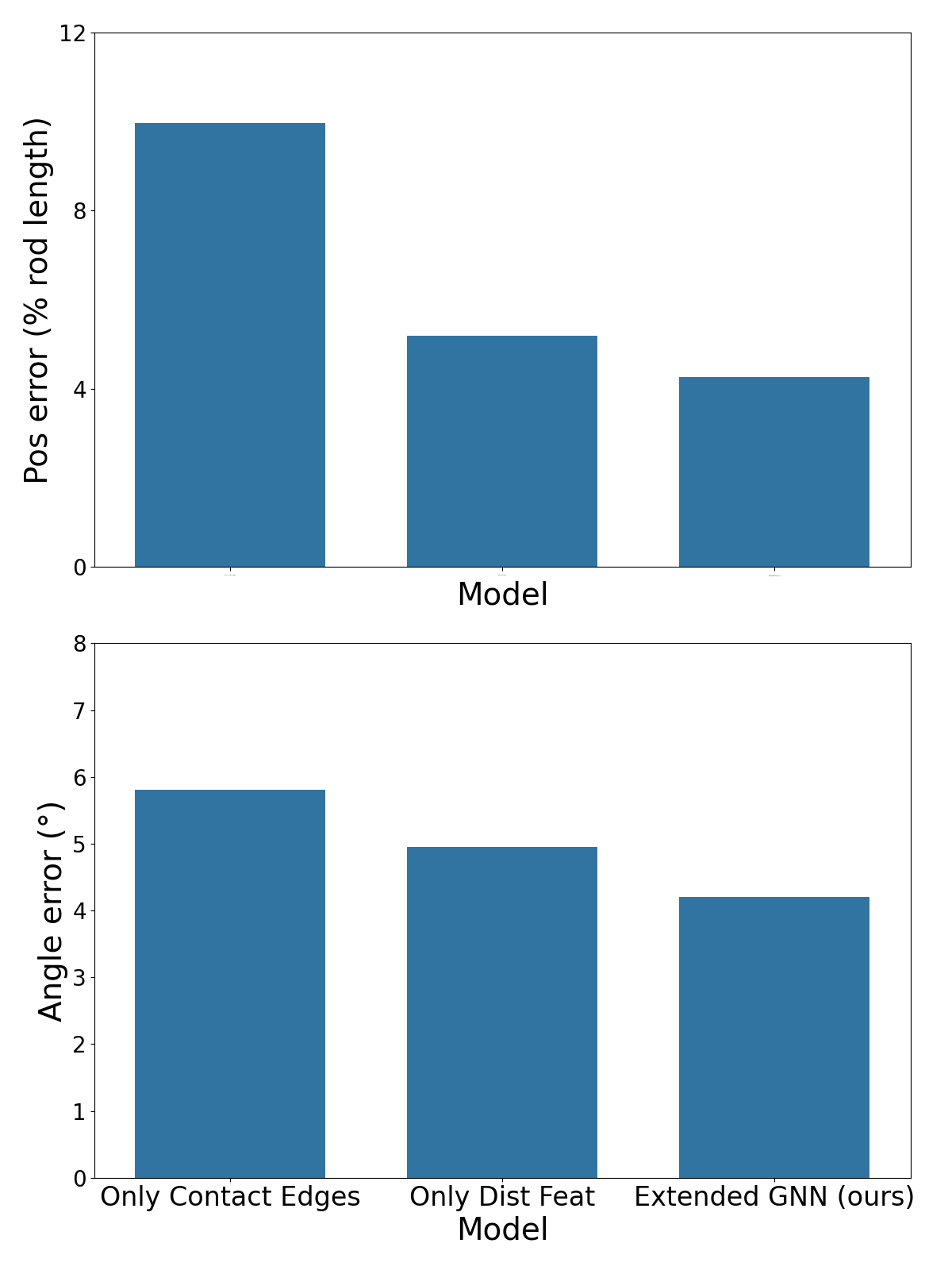}
    \vspace{-0.2in}
    \caption{An ablation on the contribution of contact edges and node distance features to model error. \textbf{LEFT}: Average mean-squared error of center-of-mass over the MPPI horizon of two seconds (200 timesteps), normalized by the rod's length. \textbf{RIGHT}: Average absolute angular error of the rods' principal-axes over the MPPI horizon of two seconds (200 timesteps).}
    \vspace{-0.2in}
    \label{fig:model_ablation}
\end{figure}

In this section, model accuracy is evaluated across successive model-training and MPPI data collection iterations. Performance is compared with the baseline GNN from \cite{chen2026cablerobotgraphsimgraphneuralnetwork}, which does not consider planar environment surfaces. An ablation study is also conducted to analyze the contribution of the proposed contact representations: (i) contact edges only, (ii) distance-to-contact node features only, and (iii) both combined (our extended GNN). Model performance is evaluated using normalized (by rod length) positional error $\mathbf{e}_{pos}$ and rotational error $\mathbf{e}_{rot}$ over the prediction horizon $H$ used by the MPPI controller. For evaluation, $N$ starting states are sampled per trajectory, then rolled out over horizon $H$, and finally, the predictions are compared against ground truth. Final errors are an average across all $K$ trajectories in the test set and are computed with

\vspace{-0.1in}
\begin{align}
\mathbf{e}_{pos}&=\frac{1}{L_{rod} NK}\sum{(P_j^{GT}-P_j^{pred})^2} \\
\mathbf{e}_{rot}&=\frac{1}{NK}\sum{\cos^{-1}{(\hat{r}_j^{GT}\cdot \hat{r}_j^{pred}})}
\end{align}

To study the impact of iterative data collection, four training iterations are performed. The initial model (iteration 0) is trained using data generated from human-designed motion primitives. In the following three iterations, an MPPI controller using the latest learned GNN model collects additional trajectories, which are then added back to the training dataset for the next model. Results are shown in Fig.~\ref{fig:model_eval_iteration}. The extended GNN consistently outperforms the baseline model, which lacks explicit information about environment contacts. Prediction errors also decrease across iterations as the dataset grows and the model is trained on trajectories generated by the controller.

The ablation study results are shown in Fig.~\ref{fig:model_ablation}. All variants are evaluated using only the iteration 0 dataset to isolate the effect of the contact representations. Both contact edges and distance-to-contact node features improve prediction accuracy. However, the distance-to-contact node feature has a larger impact on overall performance, indicating that explicit proximity information is particularly important for modeling contact dynamics.


\subsection{Navigation Tasks}


\begin{table*}[t]
\centering
\small
\begin{tabular}{c|cc|cc|cc|cc}
\toprule
 Task/Course & \multicolumn{2}{|c|}{Iteration 0} &  \multicolumn{2}{|c|}{Iteration 1} & \multicolumn{2}{|c|}{Iteration 2} & \multicolumn{2}{|c}{Iteration 3}\\
 & Success & Time & Success & Time & Success & Time & Success & Time \\
 Method & (\%) $\uparrow$ & (s) $\downarrow$ & (\%) $\uparrow$ & (s) $\downarrow$ & (\%) $\uparrow$ & (s) $\downarrow$ & (\%) $\uparrow$ & (s) $\downarrow$ \\
\midrule
\textbf{Flat Obstacle Course}& & & & & & & & \\
$A^*$ + Grid Wave-Front Heuristic& 0\% & 1200s& 73\% & 1068s& 90\% & 817s & \textbf{100}\%& 741s\\ 
$A^*$ + Reverse Motion Primitive Heuristic& \textbf{7\%}& \textbf{1184s}& 33\%& 1095s& 97\%& \textbf{432s}& \textbf{100\%}& \textbf{452s}\\
MPPI only& 0\%& 1200s& 30\%& 1144s& 47\%& 1061s& 37\%& 1087s\\
Hybrid MPPI& 0\%& 1200s& \textbf{100\%}& \textbf{701s}& \textbf{100\%}& 694s& \textbf{100\%}& 650s\\
\midrule
\textbf{Incline}& & & & & & & & \\
$A^*$ + Grid Wave-Front Heuristic& 0\%& 600s& 0\%& 600s& 0\%& 600s& 0\%& 600s\\
$A^*$ + Reverse Motion Primitive Heuristic& 0\%& 600s& 0\%& 600s& 0\%& 600s& 0\%& 600s\\
MPPI only& 0\%& 600s& 23\%& 504s& 30\%& 312s& 33\%& 300s\\
Hybrid MPPI& \textbf{100\%}& \textbf{282s}& \textbf{100\%}& \textbf{223s}& \textbf{100\%}& \textbf{274s}& \textbf{100\%}& \textbf{193s}\\
\midrule
\textbf{Narrow Corner \& Passageway}& & & & & & & & \\
MPPI only& 0\%& 400s& 3\%& 398s& 20\%& 392s& 30\%& 380s\\
Hybrid MPPI& \textbf{70\%}& \textbf{313s}& \textbf{63\%}& \textbf{323s}& \textbf{77\%}& \textbf{311s}& \textbf{80\%}& \textbf{310s}\\
\midrule
\textbf{Low Clearance Structure}& & & & & & & & \\
MPPI only& 0\%& 120s& 10\%& 117s& 37\%& 102s& 73\%& 78s\\
Hybrid MPPI& \textbf{67\%}& \textbf{97s}& \textbf{73\%}& \textbf{82s}& \textbf{60\%}& \textbf{101s}& \textbf{77\%}& \textbf{71s}\\
\midrule
\textbf{3D Obstacle Course}& & & & & & & & \\
MPPI only& 0\%& 900s& 0\%& 900s& 0\%& 900s& 35\%& 841s\\
Hybrid MPPI& \textbf{27\%}& \textbf{868s}& \textbf{25\%}& \textbf{882s}& \textbf{80\%}& \textbf{813s}& \textbf{84\%}& \textbf{747s}\\

\bottomrule
\end{tabular}
\caption{Navigation results for the three bar tensegrity robot over 5 different courses. The model training and controller execution data collection loop is executed for four iterations. At each iteration, the previous version of the model is used by the controller to execute paths for additional data generation and evaluation. Each controller is executed 30 times per course and per iteration. Success rates and average times are recorded per course and per iteration.}
\label{tab:navigation_results}
\vspace{-0.25in}
\end{table*}

The MPPI controller is evaluated on five navigation tasks shown in Fig.~\ref{fig:obs_course_1} and Fig.~\ref{fig:obs_courses_2}: (i) a flat obstacle course, (ii) a long $5^\circ$ incline, (iii) a narrow corner and passageway, (iv) a low clearance structure, and (v) a complex 3D obstacle course containing inclines, narrow passages, and a low-clearance structure. The final course is not included in the training environments and therefore represents an unseen scenario for the internal GNN dynamics model. In each task, the objective is to navigate from a start position to a goal position within a fixed time limit. A trial is considered successful if the robot reaches the goal within this limit. Each task is executed 30 times, and the success rate and average completion time are recorded.

$A^*$ with re-planning~\cite{our_ral} is used as a baseline. Two heuristics are evaluated: an obstacle-aware, grid-based wavefront heuristic and an obstacle-aware, reverse motion-primitive wavefront heuristic. The standard grid-based wavefront heuristic captures obstacle-induced shortest-paths, while the reverse motion-primitive variant additionally incorporates the robot’s motion constraints and more accurately reflects the kinematics of the 3-bar tensegrity robot.

Both $A^*$ baselines use a set of human-engineered motion primitives, including forward rolling and clockwise and counterclockwise turns. The corresponding $SE(2)$ transformations are precomputed using the GNN dynamics model and used during planning. The $A^*$ planner is evaluated only on the flat obstacle and incline environments. In the remaining three tasks, successful traversal requires intentional contact with obstacles; since the $A^*$ planner enforces collision avoidance, it cannot produce any feasible plans and is therefore not evaluated.

Similar to the GNN dynamics model evaluation, the data collection loop is run for four iterations, where the 0th iteration is based on motion primitives trajectories, and subsequent iterations generate trajectory data using the current MPPI controller. For evaluation, each controller at every iteration and environment is executed 30 times, and the success rate and average completion time are reported.

\textbf{Flat obstacle course} (bottom of Fig.~\ref{fig:obs_course_1}) contains two walls that block a direct path to the goal, requiring an S-shaped trajectory. The time limit is 1200 seconds. As shown in Table~\ref{tab:navigation_results}, both $A^*$ re-planning baselines perform well and improve across iterations as the dynamics model becomes more accurate, ultimately achieving a 100\% success rate with shorter completion times. In this task, the $A^*$ variant using the reverse-motion primitive wave-front heuristic achieves the fastest completion times. In contrast, the MPPI-only controller performs inconsistently, likely due to difficulty sampling turning maneuvers. The Hybrid MPPI approach performs poorly in iteration 0 but improves rapidly after incorporating MPPI-generated data, achieving a consistent 100\% success rate from iteration 1 onward while continuing to reduce completion time.

\textbf{Incline course} (middle of Fig.~\ref{fig:obs_courses_2}) introduces a long $5^\circ$ incline, adding a vertical climbing component. This makes primitive-based planning challenging because dynamics vary with robot heading relative to the slope. The time limit is 600 seconds. The $A^*$ baselines consistently fail to reach the goal at the top of the ramp. The robot begins climbing but gradually drifts in heading, after which the changed dynamics prevent recovery. The MPPI-only controller improves over successive iterations but never achieves high success rates. In contrast, the Hybrid MPPI variant achieves a 100\% success rate from the first iteration and continues to reduce completion time with additional training.

\textbf{Narrow corner and passageway} (left of Fig.~\ref{fig:obs_courses_2}) contains a tight corner and narrow passage that cannot be traversed using a collision-avoidant strategy such as the $A^*$ baselines. This scenario highlights MPPI's ability to intentionally interact with obstacles during planning. The time limit is 400 seconds. As in the incline task, the MPPI-only controller improves with additional data but remains relatively weak. The Hybrid MPPI variant achieves substantially higher, though not perfect, success rates.

\textbf{Low clearance structure} (right of Fig.~\ref{fig:obs_courses_2}) features a hanging obstacle that creates an opening smaller than the robot's resting height. Successful traversal requires discovering control sequences that temporarily compress the robot while moving forward. Because the primitives used by the $A^*$ baseline assume the nominal resting height, the planner again cannot reach the goal. The time limit is 120 seconds. Both MPPI-only and Hybrid MPPI achieve similar performance by iteration 3, but the Hybrid MPPI controller reaches this level one iteration earlier.

\textbf{3D obstacle course} (top of Fig.~\ref{fig:obs_course_1}) combines elements from the previous tasks, including inclines, walls, narrow passages, and low-clearance structures. This environment is excluded from the training data and therefore evaluates generalization of the learned dynamics model. The time limit is 900 seconds. The MPPI-only controller struggles to complete the course within the allotted time because of the large number of turns, remaining at a 0\% success rate until the final model iteration, where it reaches 35\%. The Hybrid MPPI strategy begins with relatively low success rates in iterations 0 and 1 but improves rapidly once the learned dynamics model becomes sufficiently accurate.

\section{CONCLUSION}
This paper presented a GNN-based MPPI controller for a three-bar tensegrity robot navigation over non-horizontal planar terrains and obstacles. The core contribution is an extended GNN dynamics model augmented with a differentiable contact detection module, enabling contact-aware prediction across ramps, walls, low-clearance structures, and rod self-collisions under partial observability. Paired with an obstacle-aware grid-based wavefront cost function and a hybrid MPPI variant that supplements learned control with human-engineered turning primitives. The system is iteratively improved through a coupled data collection and model training loop. Evaluations across five MuJoCo navigation tasks demonstrated that hybrid MPPI consistently outperforms both standalone MPPI and $A^*$ re-planning baselines, with performance improving across training iterations.

Despite these promising results, limitations remain. The current approach is restricted to planar surface representations, and extending it to fully unstructured terrains is a natural next step. The turning primitives in the hybrid MPPI are hand-engineered, and learning these directly from data could improve generality and reduce manual design effort. Future work could also explore more expressive cost functions that better capture the tensegrity's nonholonomic constraints, as well as scaling the approach to more complex multi-room or outdoor environments. Lastly, demonstrating this iterative improvement of control and modeling on a real three bar tensegrity is also planned as future work.

\section*{Acknowledgements}
\noindent \footnotesize Nelson Chen and Mridul Aanjaneya were supported in part by the National Science Foundation (NSF) under awards IIS-2132972, IIS-2238955, CCF-2312220, and IIS-2551479 as well as a research gift from Red Hat, Inc. Kostas Bekris was supported in part by NSF award IIS-1956027. Zachary Brei and Rebecca Kramer-Bottiglio were supported in part by the NSF under award CMMI-2427947. Findings reported in this work are those of the authors and do not reflect the opinions of the sponsors. Kostas Bekris is an Amazon Scholar. This work was conducted as part of his role at Rutgers University and does not necessarily represent the views of Amazon.


\bibliographystyle{IEEEtran}
\bibliography{references}


\end{document}